\pdfoutput=1

\documentclass[11pt]{article}

\usepackage[]{ACL2023}

\usepackage{times}
\usepackage{graphicx}
\usepackage{latexsym}
\usepackage{amsmath} 
\usepackage{algorithm}
\usepackage{algpseudocode}

\usepackage[T1]{fontenc}

\usepackage[utf8]{inputenc}

\usepackage{microtype}

\usepackage{inconsolata}

\usepackage{todonotes}

\newcommand{\modelname}[1]{\texttt{DictaBERT-Parse#1}}
\newcommand{\shortmodelname}[1]{\texttt{DB#1}}

\title{MRL Parsing Without Tears: The Case of Hebrew}

\author{Shaltiel Shmidman\textsuperscript{1,†}, Avi Shmidman\textsuperscript{1,2,‡}, Moshe Koppel\textsuperscript{1,2,†}, Reut Tsarfaty\textsuperscript{2,‡} \\
\textsuperscript{1}DICTA / Jerusalem, Israel \quad
\textsuperscript{2}Bar Ilan University / Ramat Gan, Israel \\ 
\texttt{\small \textsuperscript{†}\{shaltieltzion,moishk\}@gmail.com} \\
\texttt{\small \textsuperscript{‡}\{avi.shmidman,reut.tsarfaty\}@biu.ac.il}}

\begin{document}
\maketitle
\begin{abstract}

Syntactic parsing remains a critical tool for relation extraction and information extraction, especially in resource-scarce languages where LLMs are lacking. Yet in morphologically rich languages (MRLs), where parsers need to identify multiple lexical units in each token, existing systems suffer in latency and setup complexity. Some use a pipeline to peel away the layers: first segmentation, then morphology tagging, and then syntax parsing; however, errors in earlier layers are then propagated forward. Others use a joint architecture to evaluate all permutations at once; while this improves accuracy, it is notoriously slow. 
In contrast, and taking Hebrew as a test case, we present a new "flipped pipeline": decisions are made directly on the whole-token units by expert classifiers, each one dedicated to one specific task. The classifiers are independent of one another, and only at the end do we synthesize their predictions. This blazingly fast approach sets a new SOTA in Hebrew POS tagging and dependency parsing, while also reaching near-SOTA performance on other Hebrew NLP tasks.
Because our architecture does not rely on any language-specific resources, it can serve as a model 
to develop similar parsers for other MRLs.
\end{abstract}

\section{The Challenge of MRL Parsing}

Morphologically Rich Languages (MRLs) such as Hebrew present unique challenges for NLP, due to their complex word structures. Any given space-delimited word is likely to be comprised of multiple morphological tokens, because prepositions, conjunctions, and relativizers are often attached as prefixes, and accusatives or genitives are often expressed as suffixes. These affixed morphological tokens are not marked in any way, and in many cases the letters which start or end a given word can be realized variously as either part of the primary word or as separate morphological tokens, depending on the context. Parsers for MRLs are tasked with resolving these ambiguities.

Pre-neural parsing approaches\footnote{
In particular parsing pipelines that work according to the schema prescribed by the Universal Dependencies (UD) initiative \cite{de-marneffe-etal-2021-universal}, link: \url{https://lindat.mff.cuni.cz/services/udpipe/}. The specific steps of each pipeline differ somewhat from language to language; e.g. not all schemas prescribe segmentation.} generally involved pipelines consisting of multiple steps in sequence: segmentation of prefixes and suffixes; afterward, morphological tagging of segmented units; and, finally, syntactic parsing. However, as highlighted by \citet{tsarfaty-2006-integrated,cohen-smith-2007-joint,green-manning-2010-better,goldberg-tsarfaty-2008-single,tsarfaty-etal-2019-whats,seeker-cetinoglu-2015-graph}, this pipeline approach propagates and compounds errors from one stage to another, compromising the accuracy of the system. 

More recently, neural parsing models \cite{seker-tsarfaty-2020-pointer, levi2024truly, krishna-etal-2020-graph} have circumvented these problems by parsing the text with a single joint morpho-syntactic model. However, this latter approach suffers in performance, because it entails  consideration of comprehensive lattices detailing all permutations of all segmentation, morphological, and syntactic possibilities across the whole sentence.

Furthermore, all existing joint models for Hebrew parsing \cite{more-etal-2019-joint, tsarfaty-etal-2019-whats, seker-tsarfaty-2020-pointer, levi2024truly}, and likewise for other MRLs such as turkish \cite{seeker-cetinoglu-2015-graph} and Sanskrit \cite{krishna-etal-2020-keep} rely on an external lexicon which dictates the range of linguistic realizations for each word in the language. This creates complications for practical integration of the systems.

Finally, in order to accomplish their tasks, the aformentioned models all rely on a wide array of external dependencies, making them rather difficult to install and integrate. Industry developers who attempt to add NLP elements for under-resourced MRLs such as Hebrew routinely report that the existing models are simply too cumbersome to operate and too slow to run, and thus unsuited for real-world real-time systems.

In this work we propose a new  "flipped pipeline" approach, based on whole tokens, wherein a series of expert classifiers each make a set of independent predictions regarding the space-delimited words, and then afterward those predictions are synthesized into a single coherent and complete morpho-syntactic analysis, with the segmentations automatically inferred and generated from that analysis. This system is also designed for use without external lexica or dependencies. We assess our new approach and demonstrate that it achieves a new SOTA regarding dependency parsing and POS tagging, and near-SOTA scores on other NLP tasks.

\section{A New MRL Parsing Proposal}

As detailed in the previous section, existing MRL parsing systems suffer from several primary shortcomings. Pipeline architectures suffer from the compounded of errors from one level to the next; joint architectures entail slow computations of lattices covering all permutations; and almost all systems rely on external lexicons and other components. Is it possible to overcome these issues in an MRL parsing system? We believe it is, by utilizing whole-token prediction, a flipped pipeline, and eliminating the lexicon. We elaborate on each of these elements in the following sections.

\subsection{Whole-token Prediction}
In order to address the issue of propagated pipeline errors, we propose shifting to classifiers that predict morphological and syntactic functions on a whole-token basis, rather than on the basis of morphological segments. That is, the classifiers should related to each space-delimited token as an indivisible unit. By shifting to whole-token predictions, there is no longer any need to perform segmentation prior to morphological and syntactic analysis, thus side-stepping the situation in which segmentation errors at the initial layer prevent the subsequent layers from succeeding. 

To be sure, this is a striking departure from existing syntactic models for MRLs, all of which treat syntactic decisions as something to be decided at the level of the morphological segments. And indeed, prima facia, the traditional morphological segmentation approach is the more sensible linguistic approach to syntactic parsing. For instance, Hebrew proclitics may contain a preposition, a relativizer, a conjunction, or all of the above, and Hebrew suffixes often contain an accusative or a genitive. From a syntactic point of view, these are all very different elements with distinct functions, and thus syntax parsers for MRLs have always treated the proclitics and suffixes as separate units; indeed, this is prescribed by the Universal Dependencies standard. Yet, in practice, we propose, the syntactic roles of the proclitics and suffixes can be satisfactorily derived in a post-facto process, after the word-based syntactic analysis has been performed.

Supporting this stance, \newcite{goldman-tsarfaty-2022-morphology} demonstrate that substantial linguistic ambiguity exists regarding the determination of correct segmentations. Accordingly, we may reason, an artificial requirement to choose a single point of segmentation may be reducing the system's ability to correctly analyze the sentence, whereas a whole-token approach allows the classifier to more flexibly evaluate the syntactic dependencies of the sentence without first committing to specific sub-word segmentations.

\subsection{Flipped Pipeline}
As noted above, joint prediction architectures such as that proposed by \newcite{more-etal-2019-joint, levi2024truly} do sidestep the issue of pipelines, but they come at a high latency cost, because they require processing so many different permutations at once via an all-encompassing lattice. 

In order to avoid this setback, we propose a "flipped pipeline" approach consisting of two stages: in the first stage of the pipeline, dedicated expert classifiers each provide one type of linguistic prediction for the sentence; and in the second stage, these predictions are synthesized together into a single coherent parse of the sentence, adding in sub-token segmentations as relevant. We call this a "flipped pipeline" because it is the reverse of the traditional pipeline: instead of first segmenting the tokens and then predicting their morphological and syntactic functions on that foundation, we first predict the morphological and syntactic functions of the whole tokens, and then we figure out the ideal distribution of the segmentations when synthesizing the predictions together.

The key point here is that rather than a series of classifiers that build on one another, here each expert classifier operates independently, based solely upon the BERT embeddings of the input sentence. Only afterward are the predictions combined into a single coherent parse of the sentence. Thus, errors are not propagated, nor is it necessary to contend with heavy lattices.

\subsection{Eliminating the Lexicon}
As noted above, existing joint MRL parsing architectures typically rely on an external lexicographical resource in order to determine the possible segmentation, morphological and syntactic options for any given space-delimited token, and in order to build the lattices (for the case of a joint architecture). Prima facia, the use of a lexicon provides a substantial boost of accuracy, because it constrains the system from veering off into completely untenable interpretations for the chosen words. However, this constraint also boomerangs against the system. When there is an out-of-vocabulary word, or an interpolated word from another language (code switching), or a word used in a new and unusual sense (e.g. as part of a slang idiom), the lexicon is helpless. Indeed, many parsing papers such as \newcite{seker-tsarfaty-2020-pointer, levi2024truly} include sections detailing how much accuracy is lost when the lattices are "uninfused", that is, when they don't contain the full set of lexical entries needed to cover the effective use of the words in the sentence, and these shown empirical drops are substantial.

On this backdrop we present our third and final key suggestion for a better MRL parser: discard the lexicon. If we can train a model that does not rely on an external symbolic lexicon but on an LLM alone, then we will be able to handle foreign and out-of-vocabulary words gracefully. The possibility of dispensing with the lexicon is especially relevant today given the option of building upon a foundation of encoder models, such as the BERT model's family. BERT models are generally trained on huge corpus, including a wide range of different genres, and thus they are naturally exposed to the foreign words and phrases that typically appear within texts, whether prize-winning prose or down-to-earth social media. As a result, a syntactic parser based on BERT alone would not be thrown off by such foreign interpolations; on the contrary it would naturally leverage the BERT embeddings for the foreign words in order to parse them in a reasonable manner. Furthermore, BERT models excel at producing precise and appropriate embeddings based on the context alone, even when the word itself is masked or unknown. This means that completely novel word usages will still be handled with aplomb. Effectively, dispensing with the lexicon and using BERT-like encoder alone results in a natural propensity to handle code switching.

A further advantage of a lexicon-less architecture is that performance directly corresponds with that of the underlying LLM encoder. This means that the release of larger or more advanced LLM will immediately translate into an improvement in the accuracy of the parsing model (after rerunning the fine-tuning of the parsing classifiers), because the parser's choices are not constrained by any considerations other than the contextualized embeddings themselves. We may therefore reasonably expect the parser's accuracy to continually rise as new  LLMs for MRLs are released, or as existing  techniques are improved, e.g., with more optimal tokenizers, such as \newcite{yehezkel-pinter-2023-incorporating}.

\subsection{Introducing \modelname}

We hereby release \modelname{} to the community as a free and unrestricted tool for both academic and commercial use. The general-purpose model, balancing accuracy and feasibility, is the BERT-base model\footnote{\url{https://huggingface.co/dicta-il/dictabert-parse}}. Additionally, we release the BERT-large model\footnote{\url{https://huggingface.co/dicta-il/dictabert-large-parse}}, for highest accuracy. Finally, we release a scaled-down model, \modelname{-tiny}\footnote{\url{https://huggingface.co/dicta-il/dictabert-tiny-parse}}, which provides the identical functionality at a fraction of the memory requirements and even faster speed, and with only a slight drop in accuracy, for those who wish to integrate Hebrew text parsing into low-resource hardware.

\section{Model Implementation}

The overarching task is defined as follows: Given an input sequence of whole tokens \(x_1...x_n\), we aim to predict a tree structure connecting \(s_1...s_m\) segments with a predicted set of respective feature sets \(f_1...f_m\) for each segment. Notably, for MRLs it is often the case that \(m\geq n\).
The model starts off by feeding the sequence of whole tokens \(x_1...x_n\) into a pre-trained LLM encoder, and then feeds each contextualized embedding \(c_1...c_n\) into individual expert classifiers for each of the following tasks:\footnote{As a general rule, in cases where the whole token is broken up into multiple word pieces, we perform the predictions only on the first word piece. Other options for pooling word-pieces are of course conceivable, but empirically this method was proven successful for our task.} Dependency Tree Parsing, Lemmatization, Morphological Functions Disambiguation, Morphological Form Segmentation, and Named Entity Recognition. We then synthesize the outputs of the expert classifiers into a unified UD analysis. 
In what follows, we elaborate first on the implementation details of each of these classifiers, and then we describe our synthesis procedure (see Section \ref{subsec:ud-synthesis}).

\subsection{Dependency Tree Parsing Expert}
\label{subsec:dep-head}
This expert classifier aims to solve the Dependency Tree Parsing task: Given a sentence composed of whole tokens \(x_1...x_n\) the objective is to determine, for each whole token in a given sentence, which whole token it grammatically depends on (its head) according to the Universal Dependencies (UD) standard, and also to determine its syntactic relationship with that head.

\textbf{Architecture}. Given a set of contextualized embeddings \(c_1...c_n\) of the whole tokens in a given sentence, we employ a self-attention mechanism, predicting for each whole token the probability of any other whole token in the sentence being its dependent head. We use single-head scaled dot-product attention for computing for each token position \(i\) its dependent head \(h_i\) as follows:

\[ h_i = \arg\max_i \frac{c W_q \cdot (c W_k)^T}{\sqrt{d_{head}}}  \]

Where \(W_q\) and \(W_k\) are the query and key transformation matrices, \(c\) represents a matrix of \(c_1...c_n\) and \(d_{head}\) is the dimension of the attention head.

Following the work of \citet{kiperwasser-goldberg-2016-simple}, 
after identifying the dependent heads \(h_1...h_n\), we then predict the syntactic function of each relation by concatenating every \([c_i ; c_{h_i}]\) and feeding that into a linear classifier. During inference, we replace the \(\arg\max\) with an MST algorithm to construct a valid dependency tree before predicting the syntactic function of each relation. 

\subsection{Lemmatization Expert}
\label{subsec:lex-expert}

This expert classifier aims to solve a lemmatization task, wherein, given a whole token \(x_i\), we aim to identify the primary lemma of the whole token when it appears within a specific context \(x_1...x_n\). 

\textbf{Architecture}. For each contextualized embedding \(c_i\), we train the model's LM-head layer to predict the most likely token from within the encoder's vocabulary to serve as the lemma for the corresponding word in the sentence; the training is supervised with a lemmatized corpus. In our model, we used a BERT encoder, which was pretrained using the Masked-Language-Model (MLM) objective. We continue training the model with a similar objective, but instead of masking tokens and predicting those same tokens, we train it to predict the corresponding lemmas. For words whose lemma is not present in the BERT vocabulary, we train the model to predict a special \texttt{[BLANK]} token.

 The lemmatization task presents a challenge for lexicon-less models such as ours, especially when dealing with MRLs and their high degree of morphological fusion, such that the lemma is often not a substring of the corresponding word. Existing lemmatization models for MRLs rely on the ability to perform a lookup of any given word within a lexicon, and to thus determine the possible lemmas from which to choose. 
However, in our case, we cannot perform any such lookup. 

The key intuition with which we compensate for the lack of lexicon is that Hebrew lemmas are almost always valid words themselves, and these words generally appear with greater frequency than the corresponding inflected forms. This means that for any moderately frequent inflected form in the BERT model's vocabulary, we can expect that the underlying lemma will exist in the model's vocabulary as well. Indeed, the foundation BERT model we used has a rather large 128,000 token vocabulary, and in our tests on typical Hebrew corpora we find that this vocabularly has a 98\% coverage of the corresponding lemmas.

\subsection{Morphological Functions Expert}
\label{subsec:morph-head}

This expert classifier aims to solve a morphological task where, given a whole token \(x_i\), we aim to tag the POS and fine-grained morphological features of the whole token within a specific context \(x_1...x_n\). Specifically, the model predicts the part-of-speech of the main lexical value, as well as gender, number, person, and tense, wherever relevant. Since space-delimited tokens in MRLs contain additional information, this classifier is also tasked with identifying proclitic functions, and determining if a suffix is appended to the word, and if so, the function it serves, as well as its gender, number, and person. All of the labels are based on the UD tagging schema\footnote{The UD annotation guidelines can be found here: \url{https://universaldependencies.org/guidelines.html}}. 
A similar approach was employed by \citet{klein-tsarfaty-2020-getting} for predicting multiple tags for each token. However, our approach is tailored specifically to the morphological structure of words and the lexical nature of UD, e.g., recognizing that the appropriate proclitic tags differ from those applicable to the main lexical value itself.

\textbf{Architecture}. Given a contextualized embedding \(c_i\), we feed it through 5 separate classifiers, itemized below; examples are based on the input word \raisebox{-0.15\height}{\includegraphics[scale=0.14]{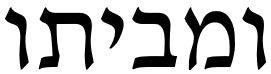}} [\texttt{"and from his house"}]:
    
\(\circ\) Prediction of the POS for the main lexical value. In our example, \texttt{house} is a NOUN. 

\(\circ\) Prediction of the proclitic functions (can be multiple or none). In our example, the word has both a CCONJ (\texttt{"and"}) and ADP (\texttt{"from"}) prefix. 

\(\circ\) Prediction of the fine-grained morphological features of the word (gender, number, person, tense). Here, \texttt{house} is singular and masculine. 

\(\circ\) Prediction of whether there is a suffix and which function it serves. In our example, there is a possessive (ADP+PRON) suffix (\texttt{"his"}). 

\(\circ\) Predictions of fine-grained features of the suffix (gender, number, and person), if the previous classifier predicted a suffix. In our example, \texttt{his} is singular, masculine, and third person. 

\subsection{Morphological Form Segmentation Expert}
\label{subsec:seg-head}

This expert classifer aims to solve a morphological task where, given a whole token \(x_i\) with a context \(x_1...x_n\), we aim to identify the actual {\em strings} that function as proclitics at the beginning of the word. 

For example, the word \raisebox{-0.15\height}{\includegraphics[scale=0.32]{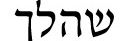}} ("that went") would be segmented into \raisebox{0.05\height}{\includegraphics[scale=0.32]{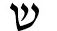}} ("that") and \raisebox{-0.15\height}{\includegraphics[scale=0.32]{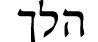}} ("went"). Note that this expert does not segment suffixes, if any, at the ends of the words (instead, suffixes are predicted by the Morphological Functions Expert classifier, detailed above).

This expert is required in addition to the expert described in Section \ref{subsec:morph-head} since Hebrew does not have a one-to-one function mapping between proclitic functions and proclitic letters. An example for this would be the implicit definite article --- a common feature of Hebrew --- which is a feature  represented via vocalization and not by a letter.

\textbf{Architecture}. Given a contextualized embedding \(c_i\), we feed it through 8 classifiers in order to predict the probability of each of the possible letter-groups which serve in a proclitic role.\footnote{For Hebrew, these are 8 types. In general the number of types for a language can be observed from the UD scheme.} During inference, we limit the predictions to valid sets of letter-groups given the initial letters of the word. 

\subsection{Named-Entity-Recognition (NER) Expert}
\label{subsec:ner-head}

This expert was trained for the named-entity-recognition task, wherein, given a sentence composed of whole tokens \(x_1...x_n\), we aim to identify and classify the named entities in the sentence. We employ the BIO tagging method, generating 27 labels from 13 classes (each class has a B and an I label, plus one O label).

\textbf{Architecture}. Given a contextualized embedding \(c_i\), a linear classifier outputs a probability vector for each of the possible labels. 

\subsection{From Expert Classifiers to a UD Tree}
\label{subsec:ud-synthesis}

\begin{figure*}
\centering
\includegraphics[width=\linewidth]{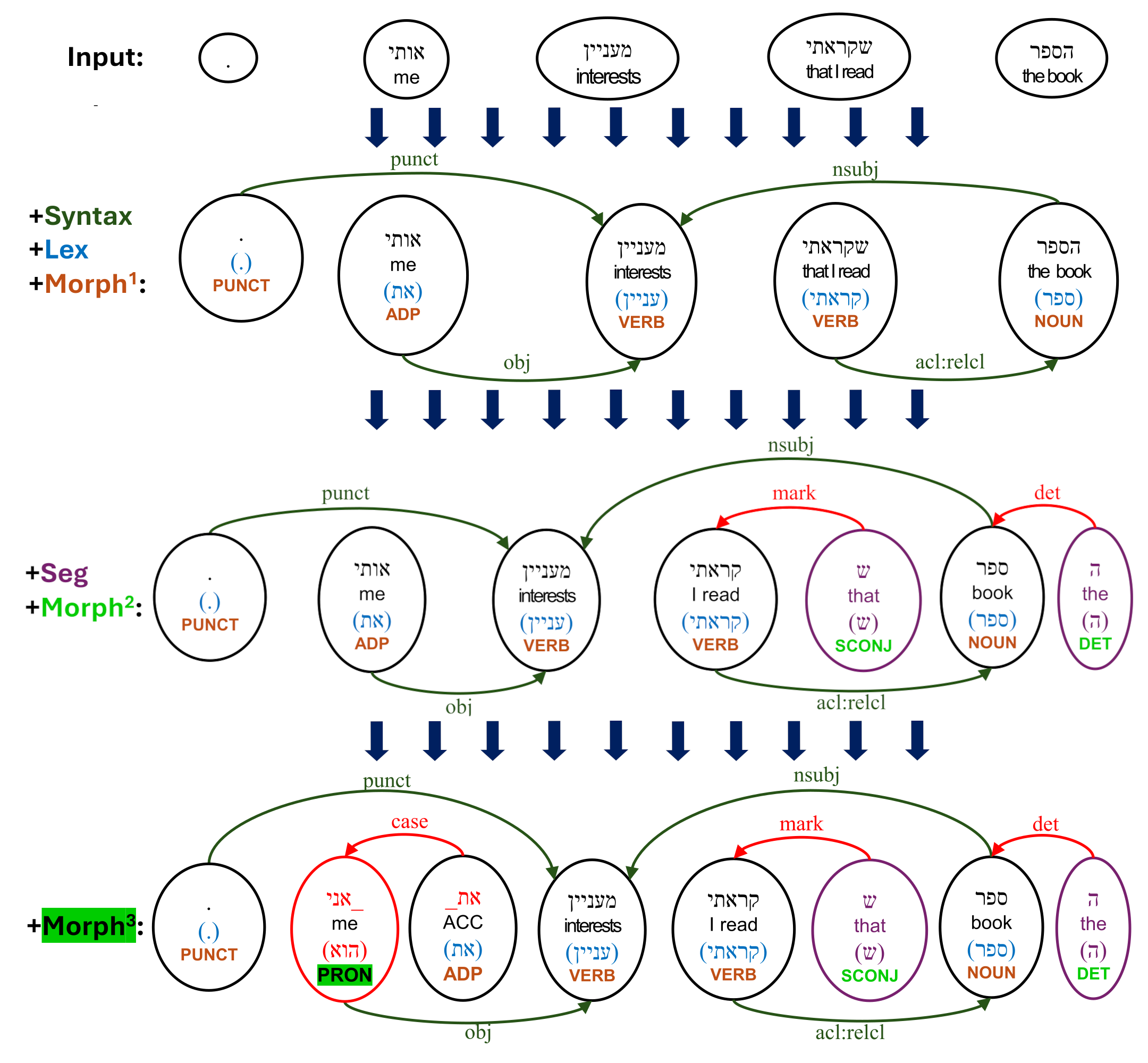}
\caption{In this figure we demonstrate the UD synthesis described in section \ref{subsec:ud-synthesis}. Hebrew is read from right to left, so the word bubbles are to be read in that order. We demonstrate a parse of the sentence "The book that I read interests me", and we present the steps by which it is broken down into its final UD analysis. In the first row, we present the initial whole-token breakdown of the sentence. The second row incorporates the labels predicted by three expert classifiers: syntax dependencies (dark green); lexemes (blue), and morphological features (Morph\textsuperscript{1}, orange). For readability, regarding the morphology classifier, we only print the POS; however, in practice the classifier predicts fine-grained morphological features as well. In the third row, we present the predictions of the expert segmentation classifier, which separates the prefixes into their own tokens (purple), paired with the output from the morphology expert (Morph\textsuperscript{2}, light green). The syntactic relations to the segmented prefixes are added automatically by the synthesis procedure (red). The bottom row demonstrates the final stage, in which we segment any suffixes into their own tokens based on the suffix features predicted by the expert morphology classifier for the corresopnding whole tokens (Morph\textsuperscript{3}, green highlight). The syntactic functions and relations of the suffixed tokens are then automatically added by a rule-based algorithm (red). }
\label{fig:ud-synthesis} 
\end{figure*}

With the flipped pipeline approach, the full UD morpho-syntactic structure is not directly available. This is because each of the expert classifiers makes predictions on a whole token basis. In order to reconstruct the full UD analysis of a given sentence, for the purpose of evaluation or a downstream application, we collect the predictions from each of the expert classifiers, and synthesize the full UD output from a combination of the predictions.  The process consists of 3 phases, which we demonstrate  in Figure~\ref{fig:ud-synthesis}. Here   we discuss the 3 phases in turn.

We start by attaching the predicted labels to the respective whole tokens. First, we take the predictions from the syntax classifier (dependency tree parsing) and build an initial whole-token dependency tree. We then continue with the lemma predictions, followed by the morphological predictions of the whole token. At this point, we have a parsed sentence on a whole-token basis. 

In the next step, we separate the prefix segments from the main words using the predictions from the segmentation expert classifier. We then assign the morphological properties of the segmented prefixes based on the prefix labels predicted by the morphology expert classifier for the corresponding whole token. The lemmas for the prefixes are automatically assigned to be equivalent to the letters of the segments.  We then assign the dependency relations to the segmented prefixes based upon the relations and functions of the whole tokens of the sentences, using rules curated based on UD labels.

Finally, we separate the suffix segments from the main words using the suffix labels predicted by the morphological expert classifier. Using these predictions, we automatically determine lemma and dependency relation of the segmented suffix. 

Pseudocode for this synthesis function is provided in Appendix \ref{sec:appendix-pseudocode}; full implementation of the algorithm in python is available with the huggingface model.

\begin{table*}[t]
\centering
\begin{tabular}{c|ccc|cc|cc|c}
\cline{2-8}
\multicolumn{1}{l|}{\textbf{}}                      & \multicolumn{3}{c|}{\textbf{Morphology}}                                                                          & \multicolumn{2}{l|}{\textbf{Dependency}}           & \multicolumn{2}{l|}{\textbf{Dep - No Punc}}        & \multicolumn{1}{l}{}                \\ \hline
\multicolumn{1}{|c|}{}                              & \multicolumn{1}{c|}{\textbf{Seg}}  & \multicolumn{1}{c|}{\textbf{POS}}   & \multicolumn{1}{l|}{\textbf{Features}} & \multicolumn{1}{c|}{\textbf{UAS}}  & \textbf{LAS}  & \multicolumn{1}{c|}{\textbf{UAS}}  & \textbf{LAS}  & \multicolumn{1}{c|}{\textbf{NER}}   \\ \hline
\multicolumn{1}{|c|}{\textbf{YAP}}                  & \multicolumn{1}{c|}{93.64}         & \multicolumn{1}{c|}{90.13}          & -                                      & \multicolumn{1}{c|}{75.73}         & 69.41         & \multicolumn{1}{c|}{-}             & -             & \multicolumn{1}{c|}{-}              \\ \hline
\multicolumn{1}{|c|}{\textbf{PtrNetMD}}             & \multicolumn{1}{c|}{94.74}         & \multicolumn{1}{c|}{91.3}           & -                                      & \multicolumn{1}{c|}{-}             & -             & \multicolumn{1}{c|}{-}             & -             & \multicolumn{1}{c|}{-}              \\ \hline
\multicolumn{1}{|c|}{\textbf{AlephBERT}}            & \multicolumn{1}{c|}{\textbf{98.2}} & \multicolumn{1}{c|}{96.20}          & 93.05                                  & \multicolumn{1}{c|}{-}             & -             & \multicolumn{1}{c|}{-}             & -             & \multicolumn{1}{c|}{83.62}          \\ \hline
\multicolumn{1}{|c|}{\textbf{AlephBERTGimmel}}      & \multicolumn{1}{c|}{98.09}         & \multicolumn{1}{c|}{96.22}          & \textbf{95.76}                         & \multicolumn{1}{c|}{-}             & -             & \multicolumn{1}{c|}{-}             & -             & \multicolumn{1}{c|}{\textbf{86.26}} \\ \hline
\multicolumn{1}{|c|}{\textbf{Levi-Tsarfaty}}        & \multicolumn{1}{c|}{97.71}         & \multicolumn{1}{c|}{94.41}          & -                                      & \multicolumn{1}{c|}{84.6}          & 81.4          & \multicolumn{1}{c|}{88.9} & \textbf{85.4} & \multicolumn{1}{c|}{-}              \\ \hline
\multicolumn{1}{|c|}{\textbf{mT5 - Small}}          & \multicolumn{1}{c|}{94.83}         & \multicolumn{1}{c|}{94.55}          & -                                      & \multicolumn{1}{c|}{-}             & -             & \multicolumn{1}{c|}{-}             & -             & \multicolumn{1}{c|}{66.74}          \\ \hline
\multicolumn{1}{|c|}{\textbf{Stanza}}               & \multicolumn{1}{c|}{89.51}         & \multicolumn{1}{c|}{-}              & -                                      & \multicolumn{1}{c|}{-}             & -             & \multicolumn{1}{c|}{-}             & 67.4          & \multicolumn{1}{c|}{-}              \\ \hline
\multicolumn{1}{|c|}{\textbf{Trankit}}              & \multicolumn{1}{c|}{95.2}          & \multicolumn{1}{c|}{-}              & -                                      & \multicolumn{1}{c|}{-}             & -             & \multicolumn{1}{c|}{-}             & 83.6          & \multicolumn{1}{c|}{-}              \\ \hline \noalign{\hrule height 2pt}
\multicolumn{1}{|c|}{\textbf{\modelname{-tiny}}} & \multicolumn{1}{c|}{97.18}         & \multicolumn{1}{c|}{96.62}          & 94.6                                   & \multicolumn{1}{c|}{87.2}          & 82.6          & \multicolumn{1}{c|}{86.9}          & 81.9          & \multicolumn{1}{c|}{80.3}           \\ \hline
\multicolumn{1}{|c|}{\textbf{\modelname{-base}}} & \multicolumn{1}{c|}{97.89}         & \multicolumn{1}{c|}{97.26} & 95.58                                  & \multicolumn{1}{c|}{89.1} & 84.7 & \multicolumn{1}{c|}{88.7}          & 84.1          & \multicolumn{1}{c|}{83.8}           \\ \hline
\multicolumn{1}{|c|}{\textbf{\modelname{-large}}} & \multicolumn{1}{c|}{97.88}         & \multicolumn{1}{c|}{\textbf{97.35}} & 95.46                                  & \multicolumn{1}{c|}{\textbf{89.5}} & \textbf{85.4} & \multicolumn{1}{c|}{\textbf{89.2}}          & 84.6          & \multicolumn{1}{c|}{84.1}           \\ \hline
\end{tabular}
\caption{Evaluation of our model versus previously reported scores of other Hebrew parsers, as detailed in section \ref{sec:experiments}. The Segmentation \& POS scores reported are aligned Multi-Set scores. For syntax dependencies we report both the labeled (LAS) and unlabeled (UAS) aligned Multi-Set scores, the first score (Dependency) including all of the tokens in the sentence (including punctuation), and the second score (Dep - No Punc) ignoring any arcs with punctuation. For NER, we report the token-based F1 score on all the labels.}
\label{tab:results}
\end{table*}

\section{Experiments and Results}
\label{sec:experiments}

\subsection{Foundation Model}

The foundation model that we assume in this work
is \texttt{DictaBERT} \cite{shmidman2023dictabert}, the SOTA BERT model for modern Hebrew. We fine-tune and evaluate our parsing model upon three sizes of the \texttt{DictaBERT} foundation: BERT-tiny\footnote{\url{https://huggingface.co/dicta-il/dictabert-tiny}}, BERT-base\footnote{\url{https://huggingface.co/dicta-il/dictabert-base}}, and BERT-large\footnote{\url{https://huggingface.co/dicta-il/dictabert-large}}.

\subsection{Data}

In this section we describe the training corpora used in our experiments to train each of the experts in our model. We use the UD HTB Treebank \cite{sade-etal-2018-hebrew}, the NEMO dataset presented by \citet{bareket-tsarfaty-2021-neural}, and an additional UD corpus and NER corpus from the IAHLT\footnote{We would like to express our thanks to IAHLT for this tagged corpus. For more information regarding the resources curated and made available by IAHLT, see: \url{https://github.com/IAHLT/iahlt.github. io/blob/main/index.md} \label{iahlt_thanks}}. Table \ref{tab:training-corpora} provides the size of each of the corpora, and no.\ of epochs performed on each during joint training.\footnote{The UD HTB Treebank only contains 5K sentences, and as described in Table \ref{tab:training-corpora} our model was trained on a significantly larger corpus than the UD HTB corpus. The extended corpus included a slightly different tagging methodology, and for the NER also included more categories of tags. Nevertheless, in order to achieve maximal alignment with the expectations of the UD gold corpus, after we trained our model on our full corpus, we continued training for several additional epochs only on the UD Treebank and NEMO corpus.} For hyperparameters see Appendix \ref{sec:appendix_hyperparams}.

\begin{table}[t]
\centering
\begin{tabular}{|c|c|c|c|}
\hline
\textbf{}              & \textbf{\# Sentences} & \textbf{\# Words} & \textbf{\# Epochs} \\ \hline
\textbf{Morph}    & 40K                  & 975K             & 15                \\ \hline
\textbf{Dep}      & 40K                  & 975K             & 15                \\ \hline
\textbf{Seg}  & 52K                  & 1.2M             & 20                \\ \hline
\textbf{Lex} & 180K                 & 5M               & 3                 \\ \hline
\textbf{NER}           & 112K                 & 2.8M             & 15                \\ \hline
\end{tabular}
\caption{Size of the corpora used to train each expert classifier, and no. epochs performed on each corpus - Morph (Morphological Disambiguation), Dep (Dependency Tree Parsing), Seg (Prefix Segmentation), Lex (Lemmatization), NER (Named Entity Recognition).}
\label{tab:training-corpora}
\end{table}

\subsection{Metrics and Evaluation}

We compare the success of our model to previously reported SOTA scores on each of the tasks.

For morphology, segmentation and dependency parsing, we report the aligned Multi-Set scores on the UD Treebank as reported by \citet{seker2021alephberta} and \citet{levi2024truly}. For dependency parsing, we report two sets of scores --- the first with all the tokens (the LAS standard Dependency Parsing Metrics), and the second ignoring any arc with punctuation (Dep - Punc). For NER, we report the whole-token based F1 score.

We compare our results to YAP\footnote{It should be noted that YAP's dependency scores were published based on the Hebrew SPMRL treebank rather than UD; the scores would presumably be a few percent higher on UD. In contrast, YAP's segmentation and POS scores were evaluated on UD in \newcite{eyal-etal-2023-multilingual}, and we report them here based on that evaluation.} \cite{more-etal-2019-joint}, PtrNetMD \cite{seker-tsarfaty-2020-pointer} AlephBERTGimmel \cite{gueta2023large}, Levi-Tsarfaty \cite{levi2024truly}\footnote{Dependency scores are reported in their paper without punctuation only. We express our thanks to the authors of the paper for collaborating with us to compute the corresponding "with punctuation" scores, as reported in the table here.}, mT5 (we report scores for the mT5-small model, which is the closest model in size to the BERT models used here) \cite{eyal-etal-2023-multilingual}, Stanza \cite{qi-etal-2020-stanza}\footnote{The scores we report for Stanza and Trankit were taken from \citet{levi2024truly}. \label{foot:stanzatrankit}} and Trankit \cite{vannguyen2021trankit}.\textsuperscript{\ref{foot:stanzatrankit}} 

\subsection{Results and Analysis}
\label{subsec:results}

Table 1 shows the performance scores for all evaluated tasks. Given the non-standard architecture of our model, with its flipped pipeline, and with its predictions on a whole-token basis, rather than on a morphological-word basis, and without any lexicon whatsoever, one might have thought that accuracy of the full-fledged tree would suffer. However, the opposite is the case. \modelname{-large} sets a new SOTA for Hebrew dependency parsing (for both UAS and LAS with punctuation, and for UAS without punctuation), and also for POS tagging. 

Significantly, even if we ignore the BERT-large, \modelname{-base} remains the highest performing model for most of these tasks; overall, we find that \modelname{-base} and even \modelname{-tiny} produce competitive scores across the board on Hebrew NLP tasks. AlephBERTGimmel remains the highest-performing model for fine-grained morphology feature prediction, while Levi-Tsarfaty maintains the highest LAS score for dependency parsing (when punctuation is ignored).

When comparing inference times between our model to the previous SOTA on dependency parsing, our model takes on average 0.0019s / 0.0016s (base and tiny, respectively) per training example, whereas the previous SOTA method \cite{levi2024truly} reported 0.257s per training example; thus our model's speed improvement is over 100x.
\subsection{New Scoring Method}
\label{subsec:new-score}

To accompany our new whole-token approach to MRL parsing, we also propose a new scoring method for morphology and dependency tree benchmarking that does not hinge on specific segmentation point choice. As demonstrated herein, the primary parsing challenge for MRLs remains the whole-token unit, whereas predictions regarding proclitics and suffixes can almost always be derived post-facto from the whole-token predictions. Therefore, we propose computing benchmarks on a whole-token basis, without separately evaluating the properties predicted to each segmented proclitic/suffix. We demonstrate the application of this new ``whole-token'' 
scoring method to our evaluations of POS scores and dependency trees.

\textbf{POS scores}: We evaluate the POS assigned to the primary segment of each word. We compute the Macro-F1 score and the accuracy score. 

\textbf{Dependency Tree scores}: We compute the LAS and UAS scores on whole tokens, before the morphological form segmentation. Since this enforces that the predicted tree will have the same number of arcs as the gold tree, we don't need to compute the aligned MultiSet score and instead just compute accuracy (as is done in other MRL parsing studies).

A key advantage of this scoring method is that accuracy on any given task is independent from all other tasks, thus obviating the need for oracle-based evaluations, as in \newcite{levi2024truly}.

We present results using this scoring method for POS and Dependency  parsing are shown in Table~\ref{tab:new-results}. In order to compare to Levi \& Tsarfaty using this method, we evaluate their output as follows: We group each set of segmented tokens into groups according to the whole tokens in the gold corpus (the segmentation process only breaks up tokens, and never combines two separate whole tokens). We consider the prediction for any given word as correct if any of the sub-tokens of a whole token point to any of the sub-tokens of the corresponding head word in the gold tree.

We see that on this granularity, our whole token approach achieves superior results; this should ultimately result in less error propagation from the parser to downstream applications that leverage these structures.

\begin{table}
\centering
\begin{tabular}{c|cl|cc|}
\cline{2-5}
\textbf{}                                    & \multicolumn{2}{c|}{\textbf{POS}}                      & \multicolumn{2}{c|}{\textbf{Dependency}}           \\ \hline
\multicolumn{1}{|c|}{\textbf{}}              & \multicolumn{1}{c|}{\textbf{Macro-F1}} & \textbf{Acc}  & \multicolumn{1}{c|}{\textbf{UAS}}  & \textbf{LAS}  \\ \hline
\multicolumn{1}{|c|}{\textbf{Levi-Tsa}} & \multicolumn{1}{c|}{88.2}              & 94.74          & \multicolumn{1}{c|}{82.1}          & 77.5          \\ \hline
\multicolumn{1}{|c|}{\textbf{\shortmodelname{-tiny}}} & \multicolumn{1}{c|}{91.7}              & 96.8          & \multicolumn{1}{c|}{89.7}          & 85.2          \\ \hline
\multicolumn{1}{|c|}{\textbf{\shortmodelname{-base}}} & \multicolumn{1}{c|}{\textbf{93.5}}     & \textbf{97.2} & \multicolumn{1}{c|}{91.3} & 87.5 \\ \hline
\multicolumn{1}{|c|}{\textbf{\shortmodelname{-large}}} & \multicolumn{1}{c|}{93.1}     & 97.0 & \multicolumn{1}{c|}{\textbf{91.6}} & \textbf{87.9} \\ \hline
\end{tabular}
\caption{Results using the new "whole-token" scoring method; "Levi-Tsa" = \newcite{levi2024truly}}
\label{tab:new-results}
\end{table}

\section{Conclusion}
In this paper we have proposed a new solution for MRL parsing, to overcome the deficiencies of current approaches (pipelines with error propagation, joint with long latencies, and cumbersome setup and integration). Our key innovation is a "flipped pipeline", in which the multiple layers involved in MRL parsing are predicted independently on a whole-token basis, and then later synthesized. Our architecture provides a substantial boost in usability over previous parsers, both in terms of speed and in terms of ease of installation and integration, while also setting a new SOTA for Hebrew POS tagging and dependency parsing. Our architecture does not rely on lexicons nor on any other language-specific linguistic resources, paving the way for it to be adapted to other MRLs as well. We release our new  parsing models to the NLP community on huggingface, under a CC BY 4.0 license.

\section{Limitations}
One of the primary achievements of this "Without Tears" project was the release of a morphosyntactic parser and lemmatizer within a single huggingface module, without the need for any external linguistic resources, including lexicons. Inherent within this, however, is a substantial limitation regarding the model's ability to predict lemmas, because it has no mechanism to predict lemmas that are not found with the vocabulary of the underlying BERT model. In practice, the model is still able to accurately predict lemmas for most words, because we use a BERT model with a fairly large vocabulary (128K), and because, following Zipf's law, the overwhelming majority of words that appear in a text tend to be drawn from the same pool of frequent words whose lemmas are covered by the BERT vocabulary. Nevertheless, when it comes to less frequent words - words which lexicon-based parsers can easily lemmatize based on a single lexicon lookup - the present model is likely to falter.

\section{Ethics Statement}
The foundation of the parser released herein is a BERT model which was trained on a large corpus of Hebrew text. The BERT model was trained on the textual corpus as is, without any editing, filtering, or censoring. This means that the parser may absorbed any biases present within the corpus itself. This issue is especially relevant for Hebrew parsers when it comes to issues of gender bias. In Hebrew, verbs generally have two different words for the masculine and the feminine, yet in practice the two words will be written with the same sequence of letters (the difference between the words is indicated via the diacritics; yet the diacritics are generally omitted in written Hebrew texts). Thus, as part of the morphological tagging task, Hebrew parsers must predict whether these ambiguous written words are in fact masculine or feminine forms, given the context. In such cases, our parser is liable to make decisions based upon stereotypical gender roles, to the extent that these roles are reflected by the texts in the corpus.

\section*{Acknowledgements}
The work of the second author has been supported by ISF grant 2617/22.
The work of the last author has been funded by the Israeli Ministry of Science and Technology (MOST) grant No.\ 3-17992, and by an Israeli
Innovation Authority  (IIA) KAMIN grant, for which we are grateful.

\bibliography{anthology,custom}
\bibliographystyle{acl_natbib}

\newpage

\appendix

\section{Appendix: Hyperparameters}
\label{sec:appendix_hyperparams}

The expert classifiers of our model are all linear classifiers, and thus their size is predetermined, since they simply map from the hidden dimension size of the base BERT model to a set number of labels. 

For the Dependency Tree Parsing expert, we had to choose the desired attention-head dimension. We attempted sizes 64, 128, 256, 512, and 768, and found that after 128 we no longer saw any improvement in accuracy; thus, we chose a size of 128.

We describe the list of the hyperparameters we used for training in Table \ref{tab:hyperparams}.

We trained the model on a single \textit{NVIDIA RTX 3090}. The total train time for the large model was 24.5 hours, for the base model 13 hours, and for the tiny model 4.5 hours.  

We evaluated the inference times of the models using a test set of sentences of varying lengths, ranging from 16 to 256 tokens. For a set of 4000 sentences processed on an \textit{NVIDIA RTX 4090} with a batch size of 8, the base model completed inference of all 8000 sentences in 15.4 seconds (0.0019 seconds per sentence), while the tiny model required only 12.8 seconds (0.0016 seconds per sentence). When processing a smaller batch of 500 sentences on a 32-core CPU with a batch size of 1, the base model completed inference of all 500 sentences in 35.4 seconds, compared to 24.3 seconds for the tiny model.

\begin{table}[h]
\centering
\begin{tabular}{|c|c|}
\hline
\textbf{Learning Rate}              & 5e-6 (5e-5 for tiny) \\ \hline
\textbf{Optimizer}                  & AdamW                         \\ \hline
\textbf{Warmup Steps}               & 5000                          \\ \hline
\textbf{Batch Size}                 & 8                             \\ \hline
\textbf{Syntax Head Size} & 128                           \\ \hline
\end{tabular}
\caption{Description of the hyperparameters used when training our model.}
\label{tab:hyperparams}
\end{table}

\newpage
\onecolumn

\section{Appendix: Pseudocode for Synthesis of Expert Classifiers into Unified UD Tree}
\label{sec:appendix-pseudocode}

Below is pseudocode for our process which synthesizes the predictions from the multiple expert classifiers into a single UD Tree. We omit the output from the NER expert, since the UD tree doesn't integrate named entities. 

The input to the function is a list of whole tokens, and the predictions for each token from the various experts: \texttt{deps} (see \ref{subsec:dep-head}), \texttt{morphs} (see \ref{subsec:morph-head}), \texttt{segs} (see \ref{subsec:seg-head}), \texttt{lemmas} (see \ref{subsec:lex-expert}).

\noindent\makebox[\linewidth]{\rule{\textwidth}{0.4pt}}
\begin{algorithmic}[1]
\Function{convertOutputToUD}{$tokens$, $deps$, $morphs$, $segs$, $lemmas$}
    \State $output \gets []$

    \For{$i = 1$ \textbf{to} $\text{length}(\textit{tokens})$}
        \State $token \gets tokens[i]$

        \State $procliticFunctions \gets morphs[i].procliticFunctions$
        \For{each $prefix$ in $segs[i]$}
            \State $pos \gets \Call{ChoosePrefixFunction}{prefix, procliticFunctions}$

            \State $depHead \gets tokens[i]$
            \If {$prefix.endswith(\text{``shin''})$}
                \State $depFunc \gets \text{``mark''}$
                \If {$morphs[i].pos \neq \text{``VERB''}$}
                    \State $depHead \gets deps[i].head$
                \EndIf
            \ElsIf {$prefix = \text{``vuv''}$}
                \State $depFunc \gets \text{``cc''}$
                \If {$deps[i].func \notin \{\text{`conj',`acl:recl', ..., (see note 1)}\}$}
                    \State $depHead \gets deps[i].head$
                \EndIf
            \Else
                \State $depFunc \gets \text{``case''}$
                \If {$morphs[i].pos \notin \{\text{\scriptsize `ADJ',`NOUN',`PROPN',`PRON',`VERB'}\}$}
                    \If {$deps[i].func \in \{\text{`aux', `det', ..., (see note 2)}\}$}
                        \State $depHead \gets deps[i].head$
                    \EndIf
                \EndIf
                \If {$prefix = \text{``heh''} \textbf{ and } pos = \text{``DET''}$}
                    \State $depFunc \gets \text{``det''}$
                \EndIf
            \EndIf
            
            \State $lemma \gets prefix$
            \State $features \gets \text{``''}$
            \State $output.append(\{(prefix, lemma, pos, features, depHead, depFunc)\})$
            \State $token \gets token[\text{length}(prefix):]$
        \EndFor
        
        \If {$\text{``heh''} \notin segs[i] \textbf { and } \text{``DET''} \in procliticFunctions$}
            \State $output.append(\{\text{Implicit Heh Entry}\})$
        \EndIf

        \State $output.append(\{token, lemmas[i], morphs[i].pos, morph[i].features, deps[i]\})$

        \If {morphs[i].hasSuffix}
            \State $output[-1].token \gets lemmas[i]$
            
            \State $\{suffix, lemma\} \gets \Call{GetSuffixToken}{morph[i].suffixFeatures}$
            \State $pos \gets morphs[i].suffixPos$
            \State $features \gets morphs[i].suffixFeatures$
            
            \State $depHead \gets tokens[i]$
            \If {$morphs[i].pos \in \{`ADP',`NUM',`DET'\}$}
                \State $depFunc \gets deps[i].func$
                \State $depHead \gets deps[i].head$
                \State $output[-1].depHead \gets suffix$
                \State $output[-1].depFunc \gets \text{``case''}$
            \ElsIf{$morphs[i].pos = \text{``VERB''}$}
                \State $depFunc \gets \text{``obj''}$
            \Else
                \State $depFunc \gets \text{``nmod:poss''}$
                \State $output.append(\{\text{Possesive Entry}\})$
            \EndIf
            
            \State $output.append(\{suffix, lemma, pos, features, depHead, depFunc\})$
        \EndIf
    \EndFor
    \State \Return $output$
\EndFunction
\end{algorithmic}
\noindent\makebox[\linewidth]{\rule{\textwidth}{0.4pt}}

\textsuperscript{1}The full list is \texttt{["conj", "acl:recl", "parataxis", "root", "acl", "amod", "list", "appos", "dep", "flatccomp"]}.

\textsuperscript{2}The full list is \texttt{["compound:affix", "det", "aux", "nummod", "advmod", "dep", "cop", "mark", "fixed"]}

The function \texttt{ChoosePrefixFunction} uses a built-in table specifying for each prefix-letter-prediction the possible proclitic functions which can be assigned to it, choosing the correct one using the predictions from the morphological function disambiguation expert. 

The function \texttt{GetSuffixToken} determines the token string and lemma using a predefined dictionary mapping from the suffix features to the relevant strings. 

We should note that in this pseudocode the values we assign to the dependency head entries are symbolic, intended to explain to the reader which value it represents. The actual implementation requires careful use of indices, making sure that each index in the whole token list points to the correct index in the segmented list. 

\end{document}